\pdfoutput=1

\documentclass[11pt]{article}

\usepackage[]{naacl2021}
\aclfinalcopy
\usepackage{times}
\usepackage{latexsym}

\usepackage[T1]{fontenc}

\usepackage[utf8]{inputenc}

\usepackage{xcolor,colortbl}

\usepackage{graphicx}
\usepackage{multirow}
\usepackage{comment}
\usepackage{amsmath}
\usepackage{url}
\usepackage{amsthm}

\definecolor{white}{rgb}{0.0,0.0,0.0} 
\definecolor{blue0}{rgb}{0.93,0.95,1} 
\definecolor{blue1}{rgb}{0.85,0.89,1} 
\definecolor{blue2}{rgb}{0.78,0.84,1} 
\definecolor{blue3}{rgb}{0.71,0.78,1} 
\definecolor{blue4}{rgb}{0.49,0.62,1} 
\definecolor{blue5}{rgb}{0.42,0.56,1} 
\definecolor{blue6}{rgb}{0.35,0.51,1} 
\definecolor{blue7}{rgb}{0.20,0.40,1} 

\usepackage{microtype}

\title{HiddenCut: Simple Data Augmentation for Natural Language Understanding with Better Generalization}

\author{
 Jiaao Chen, Dinghan Shen$^1$, Weizhu Chen$^1$, Diyi Yang \\
 Georgia Institute of Technology, $^1$Microsoft Dynamics 365 AI \\
\texttt{\{jchen896,dyang888\}@gatech.edu} \\ 
\texttt{ \{dishen,wzchen\}@microsoft.com}
}

\begin{document}
\maketitle
\begin{abstract}
Fine-tuning large pre-trained models with task-specific data has achieved great success in NLP. However, it has been demonstrated that the majority of information within the self-attention networks are redundant and not utilized effectively during the fine-tuning stage. 
This leads to inferior results when generalizing the obtained models to out-of-domain distributions. To this end, we propose a simple yet effective data augmentation technique, HiddenCut, to better regularize the model and encourage it to learn more generalizable features. Specifically, contiguous spans within the hidden space are dynamically and strategically dropped during training. 
Experiments show that our HiddenCut method outperforms the state-of-the-art augmentation methods on the GLUE benchmark, and consistently exhibit superior generalization performances on out-of-distribution and challenging counterexamples. We have publicly released our code at \url{https://github.com/GT-SALT/HiddenCut}.
\end{abstract}

\section{Introduction}
Fine-tuning large-scale pre-trained language models  (PLMs) has become a dominant paradigm in the natural language processing community, achieving state-of-the-art performances in a wide range of natural language processing tasks \cite{devlin2018bert, liu2019roberta, yang2019xlnet, joshi2019spanbert, sun2019ernie, clark2019electra, lewis2019bart, bao2020unilmv2, he2020deberta, raffel2020exploring}. Despite the great success, due to the huge gap between the number of model parameters and that of task-specific data available, the majority of the information within the multi-layer self-attention networks is typically redundant and ineffectively utilized for downstream tasks \cite{guo2020parameterefficient,gordon-etal-2020-compressing,dalvi-etal-2020-analyzing}. As a result, after task-specific fine-tuning, models are very likely to overfit and make predictions based on spurious patterns \cite{Tu2020AnES,Kaushik2020Learning}, making them less generalizable to out-of-domain distributions \cite{zhu2019freelb, jiang2019smart, aghajanyan2020better}.

In order to improve the generalization abilities of over-parameterized models with limited amount of task-specific data, various regularization approaches have been proposed, such as adversarial training that injects label-preserving perturbations in the input space \cite{zhu2019freelb, liu2020adversarial, jiang2019smart}, generating augmented data via carefully-designed rules \cite{mccoy-etal-2019-right,xie2020unsupervised,andreas-2020-good, Shen2020ASB},
and annotating counterfactual examples \cite{DBLP:conf/icml/GoyalWEBPL19,Kaushik2020Learning}. Despite substantial improvements, these methods often require significant computational and memory overhead \cite{zhu2019freelb, liu2020adversarial, jiang2019smart,xie2020unsupervised} or human annotations \cite{DBLP:conf/icml/GoyalWEBPL19,Kaushik2020Learning}.

In this work, to alleviate the above issues, we rethink the simple and commonly-used regularization technique---dropout \cite{JMLR:v15:srivastava14a}---in pre-trained transformer models \cite{Vaswani2017AttentionIA}.
With multiple self-attention heads in transformers, dropout converts some hidden units to zeros in a random and independent manner.
Although PLMs have already been equipped with the dropout regularization, they still suffer from inferior performances when it comes to out-of-distribution cases \cite{Tu2020AnES,Kaushik2020Learning}. 
The underlying reasons are two-fold:
(1) the linguistic relations among words in a sentence is ignored while dropping the hidden units randomly. In reality, these masked features could be easily inferred from surrounding unmasked hidden units with the self-attention networks. Therefore, redundant information still exists and gets passed to the upper layers. (2) The standard dropout assumes that every hidden unit is equally important with the random sampling procedure, failing to characterize the different roles these features play in distinct tasks. As a result, the learned representations are not generalized enough while applied to other data and tasks.  To drop the information more effectively, \citet{Shen2020ASB} recently introduce Cutoff to remove tokens/features/spans in the input space. Even though models will not see the removed information during training, examples with large noise may be generated when key clues for predictions are completely removed from the input. 



To overcome these limitations, 
we propose a simple yet effective data augmentation method, \textbf{HiddenCut},  to regularize PLMs during the fine-tuning stage. Specifically, the approach is based on the \textit{linguistic} intuition that hidden representations of adjacent words are more likely to contain similar and redundant information. HiddenCut drops hidden units more structurally by masking the whole hidden information of \textit{contiguous spans} of tokens after every encoding layer. This would encourage models to fully utilize all the task-related information, instead of learning spurious patterns during training. To make the dropping process more efficient, we \textit{dynamically} and \textit{strategically} select the informative spans to drop by introducing an attention-based mechanism. By performing HiddenCut in the hidden space, the impact of dropped information is only mitigated rather than completely removed, avoiding injecting too much noise to the input. 
We further apply a Jensen-Shannon Divergence consistency regularization 
between the original and these augmented examples to model the consistent relations between them. 


To demonstrate the effectiveness of our methods, we conduct experiments to compare our HiddenCut with previous state-of-the-art data augmentation method on 8 natural language understanding tasks from the GLUE \cite{Wang2018GLUEAM} benchmark for in-distribution evaluations, and 5 challenging datasets that cover single-sentence tasks, similarity and paraphrase tasks and inference tasks for \textit{out-of-distribution} evaluations. We further perform ablation studies to investigate the impact of different selecting strategies on HiddenCut's effectiveness. Results show that our method consistently outperforms baselines, especially on out-of-distribution and challenging counterexamples.
To sum up, our contributions are:
\begin{itemize}\setlength\itemsep{0em}
    \item We propose a simple data augmentation method, HiddenCut, to regularize PLMs during fine-tuning  by cutting contiguous spans of representations in the hidden space.
    \item We explore and design different strategic sampling techniques to dynamically and adaptively construct the set of spans to be cut. 
    \item We demonstrate the effectiveness of HiddenCut through extensive experiments on both in-distribution and out-of-distribution datasets.
\end{itemize}







\section{Related Work}

\subsection{Adversarial Training}
Adversarial training methods usually regularize models through applying perturbations to the input or hidden space \cite{szegedy2013intriguing,goodfellow2014explaining,aleks2017deep} with additional forward-backward passes, which influence the model's predictions and confidence without changing human judgements. Adversarial-based approaches have been actively applied to various NLP tasks in order to improve models' robustness and generalization abilities, such as sentence classification \cite{Miyato2017AdversarialTM}, machine reading comprehension (MRC) \cite{wang-bansal-2018-robust} and natural language inference (NLI) tasks \cite{nie-etal-2020-adversarial}. Despite its success, adversarial training often requires extensive computation overhead to calculate the perturbation directions  \cite{shafahi2019adversarial, zhang2019you}.  In contrast, our HiddenCut adds perturbations in the hidden space in a more efficient way that does not require extra computations as the designed perturbations can be directly derived from self-attentions.

\subsection{Data Augmentation}
Another line of work to improve the model robustness is to directly design data augmentation methods to enrich the original training set such as creating syntactically-rich examples \cite{mccoy-etal-2019-right,min-etal-2020-syntactic} with specific rules, crowdsourcing counterfactual augmentation to avoid learning spurious features \cite{DBLP:conf/icml/GoyalWEBPL19,Kaushik2020Learning}, or combining examples in the dataset to increase compositional generalizabilities \cite{jia-liang-2016-data,andreas-2020-good,chen-etal-2020-mixtext,chen2020local}. However, they either require careful design \cite{mccoy-etal-2019-right,andreas-2020-good} to infer labels for generated data or extensive human annotations \cite{DBLP:conf/icml/GoyalWEBPL19,Kaushik2020Learning}, which makes them hard to generalize to different tasks/datasets. Recently \citet{Shen2020ASB} introduce a set of cutoff augmentation which directly creates partial views to augment the training in a more task-agnostic way. Inspired by these prior work, our HiddenCut aims at improving  models' generalization abilities to out-of-distribution via linguistic-informed strategically dropping spans of hidden information in transformers.


\subsection{Dropout-based Regularization}
Variations of dropout \cite{JMLR:v15:srivastava14a} have been proposed to regularize neural models by injecting noise through dropping certain information so that models do not overfit training data. However, the major efforts have been put to convolutional neural networks and trimmed for structures in images recently such as DropPath \cite{Larsson2017FractalNetUN}, DropBlock \cite{Ghiasi2018DropBlockAR}, DropCluster \cite{Chen2020DropClusterAS} and AutoDropout \cite{pham2021autodropout}. In contrast, our work takes a closer look at  transformer-based models and introduces HiddenCut for natural language understanding tasks. 
HiddenCut is closely related to DropBlock \cite{Ghiasi2018DropBlockAR}, which drops contiguous regions from a feature map. However, different from images, hidden dimensions in PLMs that contain syntactic/semantic information for NLP tasks  are more closely related (e.g., NER and POS information), and simply dropping spans of features in certain hidden dimensions might still lead to information redundancy. 


\begin{figure*}[h]
\centering
\includegraphics[width=2.0\columnwidth]{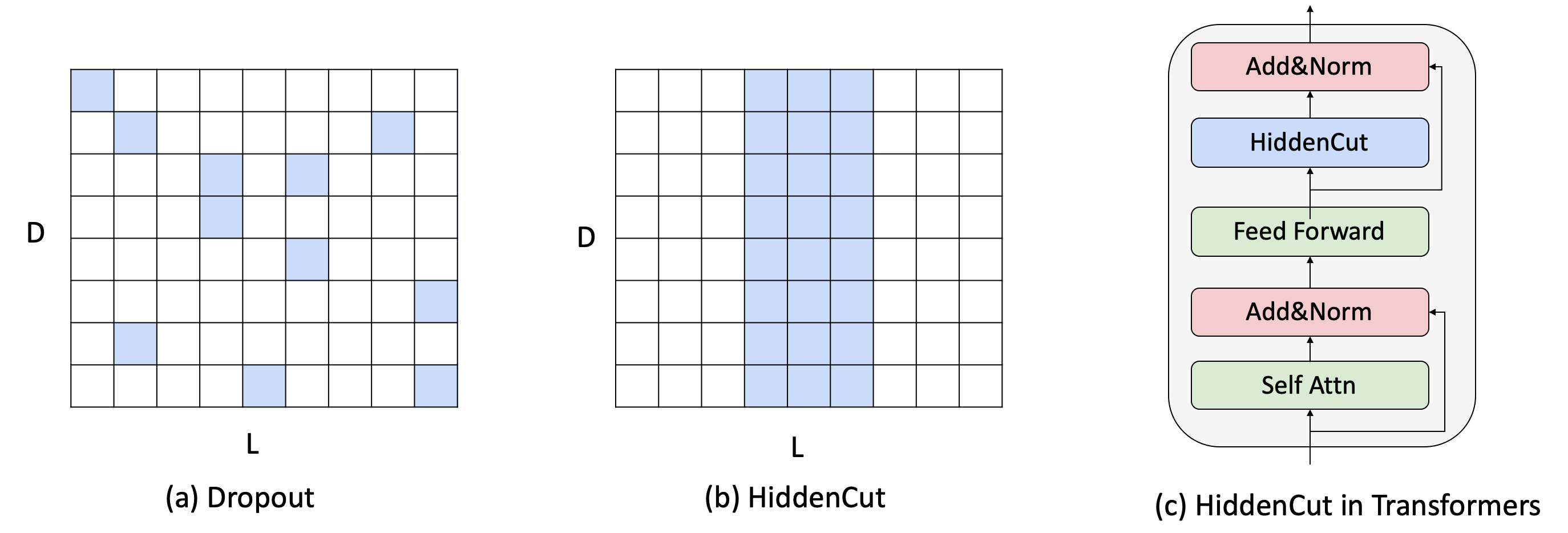}
\caption{Illustration of the differences between  Dropout (a) and  HiddenCut (b), and the position of HiddenCut in transformer layers (c). A sentence in the hidden space can be viewed as a $L \times D$ matrix where $L$ is the length of the sentence and $D$ is the number of hidden dimensions. The cells in blue represent that they are masked. Dropout masks random independent units in the matrix while our HiddenCut selects and masks a whole span of hidden representations based on attention weights received in the current layer.  In our experiments, we perform HiddenCut after the feed-forward network in every transformer layer. 
}
\label{Fig:model}
\end{figure*}

\section{HiddenCut Approach}
To regularize transformer models in a more structural and efficient manner, in this section, we introduce a simple yet effective data augmentation technique, \textbf{HiddenCut}, that reforms dropout to cutting contiguous spans of hidden representations after each transformer layer (Section~\ref{Sec:HiddenCut}). Intuitively, the proposed approach encourages the models to fully utilize all the hidden information within the self-attention networks. Furthermore, we propose an attention-based mechanism to strategically and judiciously determine the specific spans to cut (Section~\ref{Sec:Strategy}). The schematic diagram of HiddenCut, applied to the transformer architecture (and its comparison to dropout) are shown in Figure~\ref{Fig:model}.

\subsection{HiddenCut} \label{Sec:HiddenCut}






For an input sequence $\mathbf{s} = \{w_0, w_1, ..., w_L\}$ with $L$ tokens associated with a label $y$, we employ a pre-trained transformer model $f_{1:M}(\cdot)$ with $M$ layers like RoBERTa \cite{liu2019roberta} to encode the text into hidden representations. Thereafter, an inference network $g(\cdot)$ is learned on top of the pre-trained models to predict the corresponding labels. In the hidden space, after layer $m$, every word $w_i$ in the input sequence is encoded into a $D$ dimensional vector $h_i^m \in \mathrm{R}^D$ and the whole sequence could be viewed as a hidden matrix $\mathbf{H}^m \in \mathrm{R}^{L \times D}$. 

With multiple self-attention heads in the transformer layers, it is found that there is extensive redundant information across $h_i^m \in \mathbf{H}$ 
that are linguistically related \cite{dalvi-etal-2020-analyzing} (e.g., words that share similar semantic meanings). 
As a result, the removed information from the standard dropout operation may be easily inferred from the remaining unmasked hidden units. 
The resulting model might easily overfit to certain high-frequency features without utilizing all the important task-related information in the hidden space (especially when task-related data is limited). Moreover,  the model also suffers from poor generalization ability while being applied to out-of-distribution cases.

Inspired by \citet{Ghiasi2018DropBlockAR,Shen2020ASB}, we propose to improve the dropout regularization in transformer models by creating augmented training examples through HiddenCut, which drops a contiguous span of hidden information encoded in every layer, as shown in Figure~\ref{Fig:model} (c). 
Mathematically, in every layer $m$, a span of hidden vectors, $\mathbf{S}\in\mathrm{R}^{l \times D}$, with length $l = \alpha L$ in the hidden matrix $\mathbf{H}^m \in \mathrm{R}^{L \times D}$  are converted to 0, and the corresponding attention masks are adjusted to 0, where $\alpha$ is a pre-defined hyper-parameter indicating the dropping extent of HiddenCut. After being encoded and hiddencut through all the hidden layers in pre-trained encoders, augmented training data $f^{\text{HiddenCut}}(\mathbf{s})$ is created for learning the inference network $g(\cdot)$ to predict task labels.



\subsection{Strategic Sampling} \label{Sec:Strategy}


Different tasks rely on learning distinct sets of information from the input to predict the corresponding task labels. Performing HiddenCut randomly might be inefficient especially when most of the dropping happens at task-unrelated spans,  which fails to effectively regularize model to take advantage of all the task-related features. To this end, we propose to select the spans to be cut \textit{dynamically} and \textit{strategically} in every layer. In other words, we mask the most informative span of hidden representations in one layer to force models to discover other useful clues to make predictions instead of relying on a small set of spurious patterns. 

\paragraph{Attention-based Sampling Strategy} 
The most direct way is to define the set of tokens to be cut by utilizing attention weights assigned to tokens in the self-attention layers \cite{kovaleva-etal-2019-revealing}. Intuitively, we can drop the spans of hidden representations that are assigned high attentions by the transformer layers. As a result, the information redundancy is alleviated and models would be encourage to attend to other important information. Specifically, we first derive the average attention for each token, $a_i$, from the attention weights matrix $\mathbf{A} \in \mathrm{R}^{P \times L \times L}$ after self-attention layers, where $P$ is the number of attention heads and $L$ is the sequence length:
\begin{align*}
    a_i = \frac{\sum_j^P (\sum_k^L A[j][k][i])}{P}.
\end{align*}

We then sample the start token $h_i$ for HiddenCut from the set that contains top $\beta L$ tokens with higher average attention weights ($\beta$ is a pre-defined parameter). Then HiddenCut is performed to mask the hidden representations between $h_i$ and $h_{i+l}$.  Note that the salient sets are different across different layers and updated throughout the training.

\paragraph{Other Sampling Strategies} 
We also explore other widely used word importance discovery methods to find a set of tokens to be strategically cut by HiddenCut, including: 
\begin{itemize}\setlength\itemsep{1em}
    \item  \textbf{Random}: All spans of tokens are viewed as equally important, thus are randomly cut. 
    \item \textbf{LIME}  \cite{10.1145/2939672.2939778} defines the importance of tokens by examining the locally faithfulness 
    where weights of tokens are assigned by classifiers trained with sentences whose words are randomly removed.
    We utilized LIME on top of a SVM classifier to pre-define a fixed set of tokens to be cut. 
    \item \textbf{GEM} \cite{yang-etal-2019-parameter} utilizes orthogonal basis to calculate the \textit{novelty scores} that measure the new semantic meaning in tokens, \textit{significance scores} that estimate the alignment between the semantic meaning of tokens and the sentence-level meaning, and the 
    \textit{uniqueness scores} that examine the uniqueness of the semantic meaning of tokens. 
    We compute the GEM scores using the hidden representations at every layer to generate the set of tokens to be cut, which are updated during training. 
    \item \textbf{Gradient} \cite{JMLR:v11:baehrens10a}: We define the set of tokens to be cut based on the rankings of the absolute values of gradients they received at every layer in the backward-passing. This set would be updated during training.
\end{itemize}


\subsection{Objectives}
During training, for an input text sequence $\mathbf{s}$ with a label $y$, we generate $N$ augmented examples $\{f^{\text{HiddenCut}}_1(\mathbf{s}), ..., f^{\text{HiddenCut}}_N(\mathbf{s})\}$ through performing HiddenCut in pre-trained encoder $f(\cdot)$. The whole model $g(f(\cdot))$ is then trained though several objectives including general classification loss ($\mathcal{L}_{ori}$ and $\mathcal{L}_{aug}$) on data-label pairs and consistency regularization ($\mathcal{L}_{js}$) \cite{Miyato2017AdversarialTM, miyato2018virtual, Clark2018SemiSupervisedSM, xie2019unsupervised, Shen2020ASB} across different augmentations:
\begin{align*}
    \mathcal{L}_{ori} &= \mathbf{CE}(g(f(\mathbf{s})), y) \\
    \mathcal{L}_{aug} &= \sum_N \mathbf{CE}(g(f^{\text{HiddenCut}}_i(\mathbf{s})), y) \\
    \mathcal{L}_{js} &= \sum_N  \mathbf{KL}[ p(y|g(f^{\text{HiddenCut}}_i(\mathbf{s}))|| p_{avg}]
\end{align*}
where $\mathbf{CE}$ and $\mathbf{KL}$ represent the cross-entropy loss and KL-divergence respectively. $p_{avg}$ stands for the average predictions across the original text and all the augmented examples.

Combining these three losses, our overall objective function is:
\begin{align*}
    \mathcal{L} = \mathcal{L}_{ori} + \gamma \mathcal{L}_{aug} + \eta \mathcal{L}_{js} 
\end{align*}
where $\gamma$ and $\eta$ are the weights used to balance the contributions of learning from the original data and augmented data.


\begin{table*}[h]
\centering
\begin{tabular}{r|c|c|c|c|c|c|c|c|c}  \hline
\textbf{Method} & \textbf{MNLI} & \textbf{QNLI} & \textbf{QQP} & \textbf{RTE}  & \textbf{SST-2} & \textbf{MRPC} & \textbf{CoLA} & \textbf{STS-B} & \textbf{Avg}  \\ \hline \hline
RoBERTa-base   & 87.6          & 92.8          & 91.9         & 78.7          & 94.8           & 89.5        & 63.6          & 91.2           & 86.3            \\ \hline
ALUM           & 88.1          & 93.1          & \textbf{92.0}           & 80.2          & 95.3           & 90.9          & 63.6          & 91.1           & 86.8          \\ \hline
Token Cutoff   & 88.2          & 93.1          & 91.9         & 81.2          & 95.1           & 91.1          & 64.1          & 91.2           & 87.0          \\ 
Feature Cutoff & 88.2          & 93.3          & \textbf{92.0}           & 81.6          & 95.3           & 90.7          & 63.6          & 91.2           & 87.0          \\ 
Span Cutoff    & \textbf{88.4} & 93.4          & \textbf{92.0}           & 82.3          & 95.4           & 91.1          & 64.7          & 91.2           & 87.3          \\ \hline
HiddenCut $\dag$       & 88.2          & \textbf{93.7} & \textbf{92.0}  & \textbf{83.4} & \textbf{95.8}  & \textbf{92.0} & \textbf{66.2} & \textbf{91.3}  & \textbf{87.8} \\ \hline
\end{tabular} \caption{In-distribution evaluation results on the dev sets of the GLUE benchmark. $\dag$ means our proposed method.} \label{Tab: in-distribution}
\end{table*}


\begin{table*}[h]
\centering
\begin{tabular}{r|c|c|c|c|c} \hline
\multicolumn{1}{c|}{\multirow{2}{*}{\textbf{Method}}} & \multicolumn{2}{c|}{\textbf{Single-Sentence}}                       & \multicolumn{1}{c|}{\textbf{Similarity\&Paraphrase}} & \multicolumn{2}{c}{\textbf{Inference} }                        \\ \cline{2-6} 
\multicolumn{1}{c|}{}                                 & \multicolumn{1}{c|}{\textbf{IMDB-Cont.}} & \multicolumn{1}{c|}{\textbf{IMDB-CAD}} & \multicolumn{1}{c|}{\textbf{PAWS-QQP}}                       & \multicolumn{1}{c|}{\textbf{HANS}} & \multicolumn{1}{c}{\textbf{AdvNLI (A1)}} \\ \hline \hline
RoBERTa-base                                          & 84.6                            & 88.4                          & 38.4                                                 & 67.8                      & 31.2                             \\ \hline
Span Cutoff                                           & 85.5                            & 89.2                          & 38.8                                                 & 68.4                      & 31.1                             \\ \hline
HiddenCut $\dag$                                              & \textbf{87.8}                   & \textbf{90.4}                 & \textbf{41.5}                                        & \textbf{71.2}             & \textbf{32.8}  \\ \hline                  
\end{tabular} \caption{Out-of-distribution evaluation results on 5 different challenging sets. $\dag$ means our proposed method. For all the datasets, we did not use their training sets to further fine-tune the derived models from GLUE.}  \label{Tab: out-of-distribution}
\end{table*}

\section{Experiments}

\subsection{Datasets}
We conducted experiments on both in-distribution datasets and out-of-distribution datasets to demonstrate the effectiveness of our proposed HiddenCut.

\paragraph{In-Distribution Datasets}
We mainly trained and evaluated our methods on the widely-used GLUE benchmark \cite{Wang2018GLUEAM} which covers a wide range of natural language understanding tasks: \textit{single-sentence tasks} including: (i) Stanford Sentiment Treebank (SST-2) which predict the sentiment of movie reviews to be positive or negative, and (ii) Corpus of Linguistic Acceptability (CoLA) which predict whether a sentence is linguistically acceptable or not; \textit{similarity and paraphrase tasks} including (i) Quora Question Pairs (QQP) which predict whether two question are paraphrases, (ii) Semantic Textual Similarity Benchmark (STS-B) which predict the similarity ratings between two sentences, and (iii) Microsoft Research Paraphrase Corpus (MRPC) which predict whether two given sentences are semantically equivalent; \textit{inference tasks} including (i) Multi-Genre Natural Language Inference (MNLI) which classified the relationships between two sentences into entailment, contradiction, or neutral, (ii)  Question Natural Language Inference (QNLI) which predict whether a given sentence is the correct answer to a given question, and (iii) Recognizing Textual Entailment (RTE) which predict whether the entailment relation holds between two sentences. 
Accuracy was used as the evaluation metric for most of the datasets except that Matthews correlation was used for CoLA and Spearman correlation was utilized for STS-B.

\paragraph{Out-Of-Distribution Datasets}
To demonstrate the generalization abilities of our proposed methods, we directly evaluated on 5 different out-of-distribution challenging sets, using the models that are fine-tuned on GLUE benchmark datasets:
\begin{itemize}
    \item \textbf{Single Sentence Tasks}: Models fine-tuned from SST-2 are directly evaluated on two recent challenging sentiment classification datasets: IMDB Contrast Set \cite{gardner-etal-2020-evaluating} including 588 examples and IMDB Counterfactually Augmented Dataset \cite{Kaushik2020Learning} including 733 examples. Both of them were constructed by asking NLP researchers \cite{gardner-etal-2020-evaluating} or Amazon Mechanical Turkers \cite{Kaushik2020Learning} to make minor edits to examples in the original IMDB dataset \cite{maas-etal-2011-learning} so that the sentiment labels change while the major contents keep the same.
    \item \textbf{Similarity and Paraphrase Tasks}: Models fine-tuned from QQP are directly evaluated on the recently introduced challenging paraphrase dataset PAWS-QQP \cite{zhang-etal-2019-paws} that has 669 test cases. PAWS-QQP contains sentence pairs with high word overlap but different semantic meanings created via word-swapping and back-translation from the original QQP dataset. 
    \item \textbf{Inference Tasks}: Models fine-tuned from MNLI are directly evaluated on two challenging NLI sets: HANS \cite{mccoy-etal-2019-right} with 30,000 test cases  and Adversarial NLI (A1 dev sets) \cite{nie-etal-2020-adversarial} including 1,000 test cases. The former one was constructed by using syntactic rules (lexical overlap, subsequence and constituent) to generate non-entailment examples with high premise-hypothesis overlap from MNLI. The latter one was created by adversarial human-and-model-in-the-loop framework \cite{nie-etal-2020-adversarial} to create hard examples based on BERT-Large models\cite{devlin2018bert} pre-trained on SNLI \cite{bowman-etal-2015-large} and MNLI.
\end{itemize}

\subsection{Baselines}
We compare our methods with several baselines:
\begin{itemize}
    \item \textbf{RoBERTa} \cite{liu2019roberta}  is used  as our base model. Note that RoBERTa is regularized with dropout during fine-tuning.
    \item \textbf{ALUM}  \cite{liu2020adversarial} is the state-of-the-art adversarial training method for neural language models, which regularizes fine-tuning via perturbations in the embedding space.
    \item \textbf{Cutoff}  \cite{Shen2020ASB} is a recent data augmentation for natural language understanding tasks by removing information in the input space, including three variations: token cutoff, feature cutoff, and span cutoff.
\end{itemize}

\subsection{Implementation Details}
We used the RoBERTa-base model \cite{liu2019roberta} to initialize all the methods. Note that HiddenCut is agnostic to different types of pre-trained models. We followed \citet{liu2019roberta} to set the linear decay scheduler with a warmup ratio of 0.06 for training. The maximum learning rate was selected from $\{5e-6, 8e-6, 1e-5, 2e-5\}$ and the max number of training epochs was set to be either $5$ or $10$. All these hyper-parameters are shared for all the models. 
The HiddenCut ratio $\alpha$ was set 0.1 after a grid search from $\{0.05, 0.1, 0.2, 0.3, 0.4\}$. The selecting ratio $\beta$ in the important sets sampling process was set 0.4 after a grid search from  $\{0.1, 0.2, 0.4, 0.6\}$. The weights $\gamma$ and $\eta$ in our objective function were both 1. All the experiments were performed using a GeForce RTX 2080Ti.

\subsection{Results on In-Distribution Datasets}
Based on 
Table~\ref{Tab: in-distribution}, we observed that, compared to \textit{RoBERTa-base} with only dropout regularization, \textit{ALUM} with perturbations in the embedding space through adversarial training has better results on most of these GLUE tasks. However, the extra additional backward passes to determine the perturbation directions in \textit{ALUM} can bring in significantly more computational and memory overhead. 
By masking different types of input during 
training, \textit{Cutoff} increased the performances while being more computationally efficient.


In contrast to \textit{Span Cutoff}, \textit{HiddenCut} not only introduced zero additional computation cost, but also demonstrated stronger performances on 7 out of 8 GLUE tasks, especially when the size of training set is small (e.g., an increase of $1.1$ on RTE and $1.5$ on CoLA). Moreover, \textit{HiddenCut} achieved the best average result compared to previous state-of-the-art baselines. These in-distribution improvements indicated that, by strategically dropping contiguous spans in the hidden space, \textit{HiddenCut} not only helps pre-trained models utilize hidden information in a more effective way, but also injects less noise during the augmentation process compared to cutoff, e.g., \textit{Span Cutoff}  might bring in additional noises for CoLA (which aims to judge whether input sentences being linguistically acceptable or not)  when one span in the input is removed, since it might change the labels. 

\subsection{Results on Out-Of-Distribution Datasets}
To validate the better generalizability of HiddenCut, we tested our models trained on SST-2, QQP and MNLI  directly on 5 out-of-distribution/out-of-domain challenging sets 
in zero-shot settings. As mentioned earlier, these out-of-distribution sets were either constructed with in-domain/out-of-domain data and further edited by human to make them harder, or generated by rules that exploited spurious correlations such as lexical overlap, which made them challenging to most existing models. 

As shown in Table~\ref{Tab: out-of-distribution}, \textit{Span Cutoff} slightly improved the performances compared to \textit{RoBERTa} by adding extra regularizations through creating restricted input. \textit{HiddenCut} significantly outperformed both  \textit{RoBERTa}  and \textit{Span Cutoff}. For example, it outperformed \textit{Span Cutoff}. by 2.3\%(87.8\% vs. 85.5\%) on IMDB-Conts,  2.7\%(41.5\% vs. 38.8\%) on PAWS-QQP, and  2.8\%(71.2\% vs 68.4\%) on HANS consistently. These superior results demonstrated that, by dynamically and strategically dropping contiguous span of hidden representations, \textit{HiddenCut} was able to better utilize all the important task-related information which improved the model generalization to out-of-distribution and challenging adversary examples. 


\begin{table}[t]
\centering
\begin{tabular}{r|c|c} \hline
\textbf{Strategy}    & \textbf{SST-2}         & \textbf{QNLI}          \\ \hline\hline
RoBERTa & 94.8          & 92.8          \\ \hline \hline
DropBlock      & 95.4          & 93.2       \\ \hline \hline
Random      & 95.4          & 93.5          \\ \hline
LIME        & 95.2          & 93.1          \\
LIME-R      & 95.3          & 93.2          \\ \hline
GEM         & 95.5          & 93.4          \\
GEM-R       & 95.1          & 93.2          \\ \hline
Gradient    & 95.6          & 93.6          \\
Gradient-R  & 95.1          & 93.4          \\ \hline
Attention   & \textbf{95.8} & \textbf{93.7} \\
Attention-R & 94.6          & 93.4    \\ \hline      
\end{tabular} \caption{The performances on SST-2 and QNLI with different strategies when dropping information in the hidden space. 
Different sampling strategies combined with \textit{HiddenCut} are presented. ``-R'' means sampling outside the set to be cut given by these strategies.
} \label{Tab:Strategy}
\end{table}

\begin{table*}[t]
\centering
\small
\begin{tabular}{c|cccccccccc|c} \hline
\textbf{Method}     & \multicolumn{10}{c|}{\textbf{Original and Counterfactual Sentences} }                                                                         & \textbf{Prediction} \\ \hline
RoBERTa   & \textless{}s\textgreater{} & I & \cellcolor{blue0}would & rate & \cellcolor{blue3}8 & \cellcolor{blue2}stars & out & of & 10 & \cellcolor{blue4}\textless{}/s\textgreater{} & Positive          \\
HiddenCut & \textless{}s\textgreater{} & \cellcolor{blue0}I &  \cellcolor{blue0}would & rate & \cellcolor{blue2}8 & \cellcolor{blue3}stars & \cellcolor{blue0}out & \cellcolor{blue0}of & \cellcolor{blue3}10 & \cellcolor{blue0}\textless{}/s\textgreater{} & Positive          \\ \hline

RoBERTa   & \textless{}s\textgreater{} & The & movie & \cellcolor{blue1}became & \cellcolor{blue0}more & and & more & \cellcolor{blue2} intriguing   & \cellcolor{blue4} \textless{}/s\textgreater{} &  & Positive          \\
HiddenCut & \textless{}s\textgreater{} & The & movie & \cellcolor{blue0} became & \cellcolor{blue1} more & and & \cellcolor{blue1} more & \cellcolor{blue2} intriguing   & \cellcolor{blue0} \textless{}/s\textgreater{} &  & Positive          \\ 
\hline \hline
RoBERTa   & \textless{}s\textgreater{} & I & \cellcolor{blue0}would & rate & \cellcolor{blue3}8 & \cellcolor{blue2}stars & out & \cellcolor{blue0}of & 20 & \cellcolor{blue3}\textless{}/s\textgreater{} & Positive          \\
HiddenCut & \textless{}s\textgreater{} & \cellcolor{blue0}I & \cellcolor{blue0}would & rate & \cellcolor{blue3}8 & \cellcolor{blue2}stars & \cellcolor{blue0}out & of & \cellcolor{blue1}20 & \textless{}/s\textgreater{} & Negative 
\\  \hline
RoBERTa   & \textless{}s\textgreater{} & The & movie & \cellcolor{blue2} became & \cellcolor{blue0} only & slightly & more & \cellcolor{blue3} intriguing   & \cellcolor{blue4} \textless{}/s\textgreater{} &  & Positive       \\
HiddenCut & \textless{}s\textgreater{} & The & movie & \cellcolor{blue0} became & \cellcolor{blue2} only & \cellcolor{blue2} slightly & \cellcolor{blue0} more & \cellcolor{blue3} intriguing   & \textless{}/s\textgreater{} &  & Negative   \\ \hline      
\end{tabular} \caption{Visualization of the attention weights at the last layer in models. The sentences in the first section are from IMDB with positive labels and the sentences in the second section is constructed by changing ratings or diminishing via qualifiers \cite{Kaushik2020Learning} to flip their corresponding labels. Deeper blue represents that those tokens receive higher attention weights.} \label{Tab:Vis}
\end{table*}

\subsection{Ablation Studies}
This section presents our ablation studies on different sampling strategies and the effect of important hyper-parameters in \textit{HiddenCut}.

\subsubsection{Sampling Strategies in HiddenCut}
We compared different ways to cut hidden representations (\textbf{DropBlock} \cite{Ghiasi2018DropBlockAR} which randomly dropped spans in certain random hidden dimensions instead of the whole hidden space) and different sampling strategies for \textit{HiddenCut} described in Section~\ref{Sec:Strategy} (including \textbf{Random}, \textbf{LIME} \cite{10.1145/2939672.2939778}, \textbf{GEM} \cite{yang-etal-2019-parameter}, \textbf{Gradient} \cite{Yeh2019OnT}, \textbf{Attention}) based on the performances on SST-2 and QNLI. For these strategies, we also experimented with a reverse set denoted by ``-R'' where we sampled outside the important set given by  above strategies.

From  Table~\ref{Tab:Strategy}, we observed that (i) sampling from important sets resulted in better performances than random sampling. Sampling outside the defined importance sets usually led to inferior performances. These highlights the importance of strategically selecting spans to drop. (ii) Sampling from dynamic sets sampled by their probabilities often outperformed sampling from predefined fixed sets (\textit{LIME}), indicating the effectiveness of dynamically  adjusting the sampling sets during training. (iii) The \textit{attention-based} strategy outperformed all other sampling strategies, demonstrating the effectiveness of our proposed sampling strategies for \textit{HiddenCut}. (iv) Completely dropping out the spans of hidden representations generated better results than only removing certain dimensions in the hidden space, which further validated the benefit of \textit{HiddenCut} over \textit{DropBlock} in natural language understanding tasks.

\begin{table}[t]
\centering
\begin{tabular}{c|c|c|c|c|c}  \hline
$\boldsymbol{\alpha}$   & \textbf{0.05}  & \textbf{0.1}   & \textbf{0.2}   & \textbf{0.3}   & \textbf{0.4}   \\ \hline\hline
\textbf{MNLI} & 88.07 & \textbf{88.23} & 88.13 & 88.07 & 87.64 \\ \hline
\end{tabular} \caption{Performances on MNLI with different HiddenCut ratio $\alpha$, which controls the length of span to cut in the hidden space.} \label{Tab:Alpha}
\end{table}

\subsubsection{The Effect of HiddenCut Ratios}
The length of spans that are dropped by \textit{HiddenCut} is an important hyper-parameter, which is controlled by the HiddenCut ratio $\alpha$ and the length of input sentences. $\alpha$ could also be interpreted as the extent of perturbations added to the hidden space. 
We presented the results of \textit{HiddenCut} on MNLI with a set of different $\alpha$ including $\{0.05, 0.1, 0.2, 0.3, 0.4\}$  in Table~\ref{Tab:Alpha}. \textit{HiddenCut} achieved the best performance with $\alpha = 0.1$, and the performance gradually decreased with higher $\alpha$ since larger noise might be introduced when dropping more hidden information. This suggested the importance of balancing the trade-off between applying proper perturbations to regularize models and injecting potential noises.

\subsubsection{The Effect of Sampling Ratios}
The number of words that are considered important and selected by \textit{HiddenCut} is also an influential hyper-parameter controlled by the sampling ratio $\beta$ and the length of input sentences. 
As shown in Table~\ref{Tab:Beta}, we compared the performances on SST-2 by adopting different $\beta$ including $\{0.1, 0.2, 0.4, 0.6\}$. When $\beta$ is too small, the number of words in the important sets is limited, which might lead \textit{HiddenCut} to consistently drop certain hidden spans during the entire training process. The low diversities reduce the improvements over baselines. When $\beta$ is too large, the important sets might cover all the words except stop words in sentences. As a result, the \textit{Attention-based Strategy} actually became \textit{Random Sampling}, which led to lower gains over baselines. The best performance was achieved when $\beta = 0.4$, indicating a reasonable trade-off between diversities and efficiencies.

\begin{table}[t]
\centering
\begin{tabular}{c|c|c|c|c}  \hline
$\boldsymbol{\beta}$   & \textbf{0.1}  & \textbf{0.2}   & \textbf{0.4}   & \textbf{0.6}     \\ \hline\hline
\textbf{SST-2} & 95.18 & 95.30  & \textbf{95.76} & 95.46 \\ \hline
\end{tabular} \caption{Performances on SST-2 with different sampling ratio $\beta$, which controls the 
size of important token set from which HiddenCut would sample. 
} \label{Tab:Beta}
\end{table}

\subsection{Visualization of Attentions} 
To further demonstrate the effectiveness of HiddenCut, we visualize the attention weights that the special start token (``<s>'') assigns to other tokens at the last layer, via several examples and their counterfactual examples in Table~\ref{Tab:Vis}. 
We observed that \textit{RoBERTa} only assigned higher attention weights on certain tokens such as ``8 stars'', ``intriguing'' and especially the end special token ``</s>'', while largely ignored other context tokens that were also important to make the correct predictions such as scale descriptions (e.g., ``out of 10'') and qualifier words (e.g., ``more and more''). This was probably because words like ``8 stars'' and ``intriguing'' were highly correlated with positive label and \textit{RoBERTa} might overfit such patterns without probable regularization. As a result, when the scale of ratings  (e.g., from ``10'' to ``20'') or the qualifier words changed (e.g., from ``more and more'' to ``only slightly more''), \textit{RoBERTa} still predicted the label as positive even when the groundtruth is negative. With \textit{HiddenCut}, models mitigated the impact of tokens with higher attention weights and were encouraged to utilize all the related information. So the attention weights in \textit{HiddenCut} were more uniformly distributed, which helped models make the correct predictions for out-of-distribution counterfactual examples.  
Taken together,  \textit{HiddenCut} helps improve model's generalizability by facilitating it to learn from more task-related information.

\section{Conclusion}
In this work, we introduced a simple yet effective data augmentation technique, HiddenCut, to improve model robustness on a wide range of natural language understanding tasks by dropping contiguous spans of hidden representations in the hidden space directed by strategic attention-based sampling strategies. Through HiddenCut, transformer models are encouraged to make use of all the task-related information during training rather than only relying on certain spurious clues. 
Through extensive experiments on in-distribution datasets (GLUE benchmarks) and out-of-distribution datasets (challenging counterexamples), HiddenCut consistently and significantly outperformed state-of-the-art baselines, and 
demonstrated superior generalization performances.  

\section*{Acknowledgment }
We would like to thank the anonymous reviewers, and the members of Georgia Tech SALT group for their feedback. This
work is supported in part by grants from 
Amazon and Salesforce.

\bibliography{anthology,custom}
\bibliographystyle{acl_natbib}

\end{document}